# A Greedy Approximation of Bayesian Reinforcement Learning with Probably Optimistic Transition Model

Kenji Kawaguchi
BWBP Artificial Intelligence Laboratory
Tokyo, Japan
kawaguchi.kenji9@gmail.com

Mauricio Araya-López
Nancy Université / INRIA
Nancy, France
mauricio.araya@inria.fr

## ABSTRACT

Bayesian Reinforcement Learning (RL) is capable of not only incorporating domain knowledge, but also solving the exploration-exploitation dilemma in a natural way. As Bayesian RL is intractable except for special cases, previous work has proposed several approximation methods. However, these methods are usually too sensitive to parameter values, and finding an acceptable parameter setting is practically impossible in many applications. In this paper, we propose a new algorithm that greedily approximates Bayesian RL to achieve robustness in parameter space. We show that for a desired learning behavior, our proposed algorithm has a polynomial sample complexity that is lower than those of existing algorithms. We also demonstrate that the proposed algorithm naturally outperforms other existing algorithms when the prior distributions are not significantly misleading. On the other hand, the proposed algorithm cannot handle greatly misspecified priors as well as the other algorithms can. This is a natural consequence of the fact that the proposed algorithm is greedier than the other algorithms. Accordingly, we discuss a way to select an appropriate algorithm for different tasks based on the algorithms' greediness. We also introduce a new way of simplifying Bayesian planning, based on which future work would be able to derive new algorithms.

## Categories and Subject Descriptors

I.2.6 [Artificial Intelligence]: Learning

## General Terms

Algorithms, Experimentation, Theory

## Keywords

Reinforcement Learning, Uncertain Knowledge, Probabilistic Reasoning, Optimal Behavior in Polynomial Time

## 1. INTRODUCTION

Reinforcement Learning (RL) is a successful technique and has been used in a number of real-world problems [1]. RL renders us the ability to design adaptable agents that can work well in uncertain environments where the consequences of each action are not obvious (*temporal credit assignment* [2]). One remaining challenge in RL is the exploration-exploitation dilemma; agents need to *explore* the world in order to obtain new knowledge, while they must *exploit* their current knowledge to earn rewards. One elegant solution for this dilemma is Bayesian RL [3], where agents can plan to exploit possible future knowledge and hence naturally trade off between exploring and exploiting. However, except for some very limited environments, full Bayesian planning is intractable. Therefore, in general, we need to adopt some approximation techniques, such as the Monte-Carlo method [4, 5, 6, 7].

Myopic approach with *optimism in the face of uncertainty* principle is a computationally efficient way to approximate Bayesian planning. Because the intractability of Bayesian planning comes from considering all the possible future beliefs, myopic approach [5] solves the problem simply by disregarding it. To compensate the myopic way of thinking, this approach usually employs optimism to encourage agents to explore uncertain aspects of their environments. Several algorithms based on this approach have been shown to guarantee polynomial sample complexity and to work surprisingly well in practice [8, 9].

However, recent studies raised a question as to the parameter sensitivity of these algorithms. The parameters of this type of algorithm have been set to be optimal by testing the algorithms' performances with a wide range of parameter values [8, 9, 10]. However, one usually cannot tune the parameters in this way, and hence useful algorithms should work without such a thorough parameter optimization procedure [9]. Also, Brunskill [11] stated that parameter tuning is required because the number of time steps, on which these algorithms are far from optimal, is too large. That is, the algorithms' sample complexities are too large and their performances are unacceptably poor before reaching the sample numbers. In summary, it is desirable to have a fast algorithm that has less time steps with poor behaviors and maintains a high level of performance despite parameter choices.

In this paper, we propose a novel algorithm that works with a wider range of parameter values and has lower sample complexity than the previous algorithms. The proposed algorithm keeps a similar level of overall computational cost with existing fast algorithms. To present our new algorithm, we first review Bayesian RL and its approximation methods. Then, we introduce a way to effectively modify standard Bayesian planning by using the information of potentially correct MDPs. In addition, we discuss and demonstrate the proposed algorithm's properties.

## 2. BACKGROUND

In RL, the agent's goal is to maximize total returns by solving sequential decision-making problems in an environment containing some unknown aspects. The environment is represented by a Markov Decision Process (MDP) which is a tuple $\{S, A, R, P, \gamma\}$ where $S$ is a set of states, $A$ is a set of actions, $P$ is a transition probability function, $R$ is a reward function, and $\gamma$ is a discount factor. An agent takes an action in a state, which triggers a transition to another state in accordance with the transition probability function $P$, while receiving a reward based on the reward function $R$. The discount factor $\gamma$ accounts for the relative importance of immediate rewards compared to future rewards by discounting the future rewards. It also obviates the need to think ahead toward the infinite horizon. In this paper, we consider the case of discrete state space $S$ and discrete action space $A$.

To maximize the rewards received in its lifetime, an agent needs to take an action by considering the immediate consequences (i.e., immediate rewards) and possible future repercussions (i.e., rewards of the future states that the actions will lead to). In other words, most rewards are dependent on sequences of actions rather than a single action (*temporal credit assignment*). Accordingly, the agent's performance should be described by a set of actions or a policy $\pi$ that maps the state space to the action space. That is, the agent's performance can be expressed by the value of policy $V^\pi$, which is

$$V^\pi(s) = E\left[\sum_{t=0}^{\infty} \gamma^t R(s_t, \pi(s_t)) \middle| s_0 = s, \pi \right].$$

The value of a policy (or value function $V^\pi$) can be written more concretely with the Bellman's equation as

$$V^\pi(s) = R(s, \pi(s)) + \gamma \sum_{s'} P(s' | s, \pi(s)) V^\pi(s')$$

where $s'$ is the next state transitioned to from the current state. To find optimal policy $\pi^*$ and optimal value function $V^*$, instead of comparing all possible value functions $V^\pi$, it is convenient to use the Bellman's optimality equation:

$$V^*(s) = \max_a R(s,a) + \gamma \sum_{s'} P(s' | s, a) V^*(s'). \quad (1)$$

Here, the optimal policy $\pi^*(s)$ corresponds to a set of actions for each state that maximizes the right hand side of equation (1). The optimal policy $\pi^*$ or an action $\pi^*(s)$ can be found by solving equation (1) with simple algorithms such as value iteration or policy iteration [12].

In this work, we focus on model-based RL [13]. Also, we assume that the unknown aspect of the environment is the transition probability function $P$ and that the agent ought to learn it to use equation (1). In this setting, the agent needs to estimate $P$ based on observations. Maximum Likelihood Estimation (MLE) is a straightforward way to do this. However, the agent that uses equation (1) along with $P$ estimated with standard MLE would become stuck in sub-optimal policy [3]. This is because that the agent does not have any intention to explore new state-action pairs in order to gain new knowledge.

### 2.1 Bayesian Reinforcement Learning

One elegant solution for the exploration-exploitation dilemma is the Bayesian approach [3], which explicitly accounts for transitions of agents' beliefs. This means that in Bayesian planning, the agent recognizes the transitions of its belief $b$ besides the transitions of the environment's states $s$. We denote the expected value of the transition probability $P$ based on the current belief $b$ as

$$P(s' | b, s, a) \triangleq E(P(s' | s, a) | b)$$

where $b$ is determined by the initial belief $b_0$ and the agent's experience. Then, we can write the Bellman's equation for Bayesian RL as follows:

$$V^*(b,s) = \max_a R(s,a) + \gamma \sum_{s'} P(s' | b, s, a) V^*(b', s') \quad (2)$$

where $b'$ is the possible next belief when the transition $(s,a,s')$ is observed with the belief $b$. Now, the optimal policy $\pi^*(b,s)$ corresponds to a set of actions that maximize the right hand side of equation (2). As it can be seen in equation (2), when a Bayesian optimal agent chooses actions, it considers how the actions affect its knowledge as well. Hence, a Bayesian optimal agent naturally solves the trade-off between the exploration for better knowledge and the exploitation of its current knowledge.

However, computing Bayesian value function in equation (2) is usually not possible. As the number of possible belief states is typically very large, full Bayesian planning with equation (2) is intractable in most cases (one exception is the k-armed bandit problem). Therefore, some approximation techniques are required.

One straightforward approach to approximate Bayesian planning is to use the Monte Carlo method. As the Monte Carlo method has been used to deal with the curse of dimensionality in many problems (e.g. see [14]), the same approach can be used for the belief-state space of Bayesian RL. Sparse Sampling [4] is a direct application of the Monte Carlo method for dynamic programing described in the Bellman's equation of both standard RL and Bayesian RL. Since a common drawback of the Monte Carlo method is its computational time, a number of algorithms have been developed based on Sparse Sampling chiefly in order to expedite its calculation time [5, 6, 15].

In particular, Bayesian Sparse Sampling [5] was proposed for Bayesian setting by modifying Sparse Sampling. It is worth understanding the concept of this algorithm as the intuitions behind it and our proposed method have some similarities. Unlike original Sparse Sampling, which looks ahead at all possible scenarios towards certain degrees of the future, Bayesian Sparse Sampling utilizes the information embedded in the agent's belief in order to pick up possible future scenarios to be sampled. To do so, it looks ahead at scenarios led only by actions that are potentially optimal based on the belief (and a myopic strategy). Because of that, it not only effectively allocates samples in practice, but it also works in the cases of both continuous and discrete action spaces unlike the original Sparse Sampling.

### 2.2 Myopic Approach with Optimism

The algorithms using the Monte Carlo method like Sparse Sampling can guarantee near-optimal behavior in theory. However, the use of sampling in the planning phase slows down the decision-making speed. A computationally faster way to approximate Bayesian planning is myopic approach with *optimism in the face of uncertainty* principle.

In myopic approach [5], an agent does not explicitly consider the effects of its actions on future beliefs and thus it is myopic. But, the use of *optimism in the face of uncertainty* principle compensates for this myopic way of thinking. Because myopic planning is optimal if the current agent's knowledge is perfect, the agent takes actions by favoring to reduce epistemic uncertainty so that the myopic way of thinking will be justified in the end. One way to force an agent to favor uncertain states is to let the agent believe that it can get the most preferable outcomes in the range of uncertainty (*optimism in the face of uncertainty*). By doing so, some algorithms guarantee near-optimal behavior. For example, R-max [16] assumes that *unknown* states have maximum rewards, and it assures PAC-MDP behavior. On the other hand, several algorithms use the Bellman's equation with the exploration reward bonus as

$$\tilde{V}^*(b,s) = \max_a \tilde{R}(s,a) + \gamma \sum_{s'} P(s'|b,s,a)\tilde{V}^*(b,s') \quad (3)$$

where $\tilde{R}$ is defined to be reward $R$ plus exploration reward bonus $R'$. One algorithm that uses MLE version of equation (3) is Model Based Interval Estimation with Exploration Bonus (MBIE-EB) [17]. MBIE-EB ensures PAC-MDP behavior, but it does not use prior information. In order to make use of prior information, equation (3) was employed with the bonus being the posterior variance in Variance-Based Reward Bonus (VBRB) [18]. VBRB guarantees PAC-MDP behavior like R-max and MBIE-EB. However, to make sure that the behavior is (probably approximately) correct regarding true MDP, these PAC-MDP algorithms show over-exploration despite prior knowledge and may not be preferable in this sense.

Bayesian Exploration Bonus (BEB) [8], which uses equation (3), is one of the few algorithms that ensure near-Bayesian optimal policy without sampling. Another such algorithm is Bayesian Optimistic Local Transitions (BOLT), which uses modified transition probability model $\tilde{P}$ rather than reforming the reward function $R$. That is, BOLT employs

$$\tilde{V}^*(b,s) = \max_{a,\tilde{s}} R(s,a) + \gamma \sum_{s'} \tilde{P}(s'|b,s,a,\tilde{s})\tilde{V}^*(b,s') \, . \quad (4)$$

For BOLT, $\tilde{P}(s'|b,s,a,\tilde{s})$ corresponds to the transition model based on the belief modified by a certain number $\eta$ of artificial observations $\tilde{s}$. The number $\eta$ is the parameter of BOLT. For instance, for independent Dirichlet distribution per each state-action pair, known as Flat-Dirichlet-Multinomial (FDM) [6], BOLT's modified transition model can be written as follows:

$$\tilde{P}_{BOLT}(s'|b,s,a,\tilde{s}) = \frac{\alpha(s,a,s') + \eta \mathbf{1}(s',\tilde{s})}{|\alpha(s,a)| + \eta} \quad (5)$$

where $\mathbf{1}(x,y)$ is the indicator function (or Kronecker delta), which outputs 1 if $x$ and $y$ are equal, and otherwise returns 0.

BOLT has a distinct advantage over BEB or other algorithms that use equation (3). In BOLT, the degree of optimism is bounded by probabilistic law for any parameter choice; that is, the parameter $\eta$ takes place in the numerator and denominator of the right hand side of equation (5), and thus the right hand side is bounded by 1.0. Because of this, BOLT is less sensitive to its parameter values and it works well for a wider range of parameter values than BEB [9]. But still, its performance decreases a lot when the parameter is not well tuned.

## 3. ALGORITHM: PROBABLY OPTIMISTIC TRANSITION

Finding the algorithms' optimal parameter values is not possible in many practical situations, and therefore we want algorithms to work well in a wide range of parameter values. However, existing algorithms that guarantee optimality with myopic approach perform well only with a narrow range of parameter values. In this section, we propose a new algorithm, called Probably Optimistic Transition (POT), which can perform better and work in a wider range of parameter choices, compared with existing algorithms.

The main reason why existing algorithms do not generalize well for a large set of parameter values is that the degree of optimism becomes unreasonably high or low unless a perfect parameter value is assigned. The key intuition behind POT is that we can adaptively adjust the degree of optimism by combining knowledge of potentially true MDP into Bayesian planning. In a standard optimistic approach, an agent expects maximum outcomes that the agent *believes to be possible* while thinking ahead. On the other hand, POT lets an agent have the most preferable models that the agent will *actually be able to obtain with high probability* in the future. In other words, an agent with POT uses probably optimistic transition models.

Formally, along with equation (4), POT uses the following transition probability model:

$$\tilde{P}(s'|b,s,a,\tilde{s}) = E\left(P(s'|s,a) \middle| b, \tilde{h}^\theta(s,a,\tilde{s})\right)$$

where $\tilde{h}^\theta$ is the number $\theta$ of artificial observations regarding transitions from states $s$ to $\tilde{s}$ with actions $a$. For example, the transition probability model of POT can be written for FDM as

$$\tilde{P}(s'|b,s,a,\tilde{s}) = \frac{\alpha(s,a,s') + \theta(s,a,s')\mathbf{1}(s',\tilde{s})}{|\alpha(s,a)| + \theta(s,a,s')} \, .$$

Here, unlike BOLT, $\theta$ is not the parameter of the algorithm, but instead it is a function of both the parameter of POT $\beta$ and knowledge of likely true MDP. Specifically for FDMs, $\theta$ corresponds to

$$\theta(s,a,s') = \beta\left(\frac{\alpha(s,a,s')}{|\alpha(s,a)|} + \sigma(s,a,s') + \frac{1}{\sqrt{2\beta}}\right) \quad (6)$$

where $\sigma$ is the posterior variance for each transition probability. The right hand side of equation (6) represents the possible number of observing each transition based on potentially true MDP. As can be seen, decreasing the parameter $\beta$ diminishes the degree of optimism, but the last term of the right hand side penalizes non-optimistic parameter setting and balances the degree. Theoretical support of the equation is discussed later. Note that when $\theta$ is greater than time horizon $H$, POT replaces $\theta$ by $H$.

### 3.1 Theoretical Properties of the Algorithm

It is convenient to consider the computational effectiveness of this type of algorithm in two different levels: the computation time per action (or per time step) and the number of actions required before achieving the optimal learning behavior (sample complexity). For the former, a disadvantage of POT seems, at first glance, to be its calculation time due to its extra argument $\tilde{s}$, compared to BEB, and $\sigma$, in comparison with BEB and BOLT. However, remember that the optimism of POT is more tightly bounded. With fixed discount factor $\gamma$ and convergence criteria, it means that value iteration in POT converges in a smaller number of steps, which is closer to the actual horizon of Bayes-optimal planning[1]. Thus, for an agent to take an action, POT could be faster than both BEB and BOLT in practice.

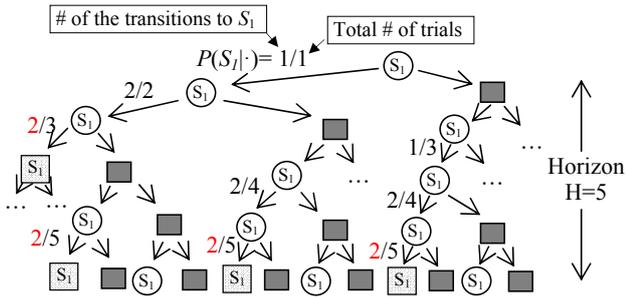

**Figure 1. A simple example of PUB Bayesian Planning (planning starts with a state $S_1$ and the agent would move to either $S_1$ (right) or other states (left) indicated by the black box by taking some actions. The numbers along with state transition arrows (e.g., 1/1, 2/2, 2/3) abstractly illustrate the agent's belief evolution of the transition probability from any states to $S_1$. The transition should happen less than 3 times with high probability in this example, and the agent does not update its belief with events violating this knowledge as indicated by red.)**

To discuss the sample complexity of POT, we first introduce a modification of Bayesian RL behavior. POT guarantees polynomial sample complexity to let an agent behave nearly as well as this modified version of Bayes-optimal learning. As we will show shortly, POT has lower sample complexity than both PAC-MDP and near Bayes-optimal algorithms.

Moreover, both the modified version of Bayes-planning and POT *exploit* additional *current knowledge*, and thus they allow agents to use greedier exploration methods.

### 3.1.1 Modification of Bayes-Optimal Planning with the Knowledge regarding Probably True MDP

In order to illustrate POT's property, we present a way of simplifying Bayesian planning by relaying on the information of likely correct MDP. Due to its characteristic, we call this simplified Bayesian planning "Probably Upper Bounded belief-based Bayesian planning", or "PUB Bayesian planning" in short.

The idea behind it is similar to the concept underling Bayesian Sparse Sampling. While Bayesian Sparse Sampling limits the possible future scenarios to be thought about based on "*myopic heuristics*" for "*likely optimal actions*", PUB Bayesian planning does so in accordance with "*probability theory*" for "*likely correct MDP*". Also, unlike Bayesian Sparse Sampling, it does *not* omit any progression in state-action space, but only unreasonable belief evolutions. Concretely, in PUB Bayesian planning, the agent does not consider belief evolutions with sets of events that should not happen with high probability. Figure 1 shows a simple example. If the transition probability from any states to $S_1$ is upper bounded by 0.01, with probability at least 0.95, the agent will observe the transition less than 3 times according to Hoeffding's inequality.

---

[1] This statement does not hold if an agent modifies $\gamma$ and convergence criteria to account for the changed scale of reward values.

The way that PUB Bayesian planning simplifies standard Bayesian planning is analogous to the following situation. We cannot identify an exact occurrence-probability of nuclear accidents. Thus, in the future, we may believe that the probability is much higher than currently believed. However, it would be impractical to assume that we will believe that the accidents happen every day in nature, even while imagining a scenario where sequential accidents coincidentally occurred over a number of days.

### 3.1.2 Sample Complexity

In this section, we show that for FDMs, POT holds optimism for PUB Bayesian RL and guarantees polynomial sample complexity for near PUB Bayes-optimal behavior.

First, we make the relationship between POT and PUB Bayesian RL clear (in the following lemma), and also reveal how POT can derive the information regarding probably true MDP by using Chebyshev's inequality (in the proof of the lemma).

**Lemma 1.** *Define $z(s,a,s')$ be the maximum number of belief updating for each transition in PUB Bayesian planning. Let $\beta$ in equation (6) be equal to $H\lambda$ where $\lambda$ is any positive real number that is at least 1. Then for each transition model, the number $\theta$ in POT is no less than $z$ with probability at least $(1-1/\lambda^2)^2$.*

*Proof.* If true transition probabilities lie at least within the belief space, we can infer the upper bounds of the values by using Chebyshev's inequality. That is, based on mean estimations and posterior variance $\sigma$,

$$P(s'|s,a) \leq \frac{\alpha(s,a,s')}{|\alpha(s,a)|} + \lambda\sigma(s,a,s')$$

with probability at least $1-1/\lambda^2$. Notice that this way to bound true MDP's values was used in [18], but unlike POT, in order to achieve PAC-MDP behavior. Then, by applying Hoeffding's inequality to the upper bound above, with probability at least $(1-1/\lambda^2)^2$, the maximum occurrence-number of transitions $z$ can be bounded as

$$z(s,a,s') \leq H\left(\frac{\alpha(s,a,s')}{|\alpha(s,a)|} + \lambda\sigma(s,a,s') + \sqrt{\frac{\ln \lambda^2}{2H}}\right) \quad (7)$$

Because of the assumptions, $\lambda \geq 1$ and $\beta \geq H\lambda$, and due to $\lambda \geq \ln\lambda^2$,

$$z(s,a,s') \leq H\lambda\left(\frac{\alpha(s,a,s')}{|\alpha(s,a)|} + \sigma(s,a,s') + \frac{1}{\sqrt{2H\lambda}}\right)$$
$$\leq \theta(s,a,s') \qquad \square$$

For instance, the maximum number of belief updating $z$ is 2 in the example illustrated by figure 1. As can be seen, the final step in the proof of lemma 1 is just to restrict the number of the free parameters that POT has. Thus, if more than one parameter in the algorithm is allowed, instead of employing equation (6), one can use equation (7) to improve its learning performance. But, in this paper, we use only equation (6).

**Lemma 2** (Optimism). *Let $V^*_{PUB}(b,s)$ and $\tilde{V}^A(b,s)$ denote the optimal PUB Bayesian value function and the value function used by POT respectively. Define $c$ to be the number of value function updates. Let the parameter $\beta$ be at least $H\lambda$. Then with probability at least $1-2|S||A|c/\lambda^2$,*

$$V^*_{PUB}(b,s) \leq \tilde{V}^{\mathcal{A}}(b,s)$$

*Proof.* Following the proof of lemma 4.1 in [9], we have

$$V^*_{PUB}(b_{i+1}, s_{i+1})$$
$$\leq \tilde{V}^{\mathcal{A}}(b_{i+1}, s_{i+1}) + \frac{\min(z(s,a,s'), i+1)f_1}{|\alpha(s,a)|+(i+1)} - \frac{\theta(s,a,s')f_1}{|\alpha(s,a)|+\theta(s,a,s')} - \gamma\Delta_i$$

where $\Delta_i$ is the positive difference between $V^*_{PUB}$ and $\tilde{V}^{\mathcal{A}}$ at $i$ step of value iteration, and $f_1$ represents some arbitrary positive value. By noticing that the second term in the right hand side of the equation above reaches its maximum when $(i+1)$ is equal to $z$, we can rewrite the inequality as follows:

$$V^*_{PUB}(b_{i+1}, s_{i+1}) \leq \tilde{V}^{\mathcal{A}}(b_{i+1}, s_{i+1}) - \frac{\theta(s,a,s') - z(s,a,s')}{|\alpha(s,a)|+\theta(s,a,s')}f_1 - \gamma\Delta_i$$
$$\leq \tilde{V}^{\mathcal{A}}(b_{i+1}, s_{i+1}) - \gamma\Delta_i$$
$$\leq \tilde{V}^{\mathcal{A}}(b_{i+1}, s_{i+1}).$$

The second line is due to the fact that $z$ is at most $\theta$ with probability at least $1-2/\lambda^2$ per transition probability. Because this needs to be true for the most preferable transition per state-action pair, the inequality holds with probability at least $1-2|S||A|/\lambda^2$. In turn, for this to be true for the entire execution of POT, it must be true for all value function updates, which results in the upper probability bound $1-2|S||A|c/\lambda^2$. The last line can be shown by induction. □

Sorg et al. [18] indicated that the number of value function updates $c$ is upper bounded by $|S||A|$, but we continue using the notation $c$ in this paper. Finally, based on the discussions so far, we can present POT's sample complexity.

**Theorem 1.** *Let $V^{\mathcal{A}}(b,s)$ denote the value function described in equation (2) with a policy of POT rather than with a Bayesian optimal policy. Suppose that the agent stops updating its belief after $|\alpha(s,a)|=4\theta^2/(\epsilon(1-\gamma))$. Let the parameter $\beta$ be at least $H\lambda$ and let $\lambda^2$ be at least $4|S||A|c/\delta$. Then, POT will follow a policy $\epsilon$-close to PUB Bayesian optimal policy with probability at least $1-\delta$ as*

$$V^*_{PUB}(b,s) \leq V^{\mathcal{A}}(b,s) - \epsilon$$

for all but

$$O\left(\frac{|S||A|\theta^2}{\epsilon^2(1-\gamma)^2}\ln\frac{2|S||A|}{\delta}\right) \leq O\left(\frac{|S||A|H^2}{\epsilon^2(1-\gamma)^2}\ln\frac{2|S||A|}{\delta}\right)$$

time steps.

*Proof.* (Sketch) Along with lemmas 1 and 2 in this paper, by directly following the procedure of the proof of theorem 5.1 in [9] with $\eta$ substituted by $\theta$, this can be easily shown. □

POT should use less time steps than that of the previous algorithms in practice in order to achieve its desired behavior. For example, that of BOLT is $\tilde{O}(|S||A|\eta^2/(\epsilon^2(1-\gamma)^2))$ where $\eta$ is greater than or equal to $\theta$. Also, the sample complexity of BEB is $\tilde{O}(|S||A|H^4/(\epsilon^2(1-\gamma)^2))$ and that of PAC-MDP algorithms are usually much greater. But, of course, we have to notice that the sample complexity of POT is for PUB Bayesian optimal policy, which is a weaker concept than standard Bayesian optimal policy. We do not discuss the exact relationship between these two types of optimal polices in this paper, leaving it as future work.

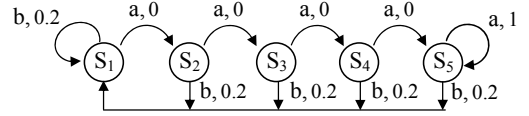

**Figure 2. Chain problem (the lower case letters 'a' and 'b' represent actions and the numbers express reward values)**

## 4. EXPERIMENT

In this section, we present the performances of POT and the other existing algorithms in the 5-state chain environment [3], which is a standard benchmark problem in the literature. Figure 2 illustrates the environment that the agent is in. In all states, the agent can choose between two actions, 'a' or 'b'. Action 'a' leads the agent to $S_{j+1}$ from $S_j$ if $j<5$, and if $j=5$ (i.e. if the agent is at $S_5$), the action lets the agent stay at $S_5$. On the other hand, action 'b' leads the agent to $S_1$ from any of the states. But, with the probability 0.2, the agent "slips" and performs the opposite action as intended. Rewards are 0.2 for returning to $S_1$, 1.0 for staying at $S_5$, and 0 otherwise. Even though the optimal policy is to always select action 'a', this setting encourages a non-exploring agent to settle on $S_1$ by taking action 'b'. We assume that the dynamics of transitions among states are completely unknown (this situation was called the "full version" of the problem in [10]). We use discount factor $\gamma=0.95$ and convergence criteria equal to 0.1 in value iteration. To make our results comparable with previously published results, we report the algorithms' performances by showing cumulative rewards in the first 1000 steps.

### 4.1 Results

In this section, we show the algorithms' performances for three different situations according to our prior knowledge about the transition probabilities. Our focus is on situations with no prior knowledge, varying degrees of useful knowledge, and different magnitudes of misleading knowledge.

First, we focus on the case where we have no previous knowledge regarding the transition probabilities. In that situation, we can assign the probability 0 for each transition probability and large rewards for the unexplored transitions as in MBIE-EB. But, for algorithms derived in Bayesian setting, we instead assume uniform prior for each transition model with a small amount of information ($\alpha(s,a,s')$=0.02) because this is a more natural approach in Bayesian statistics.

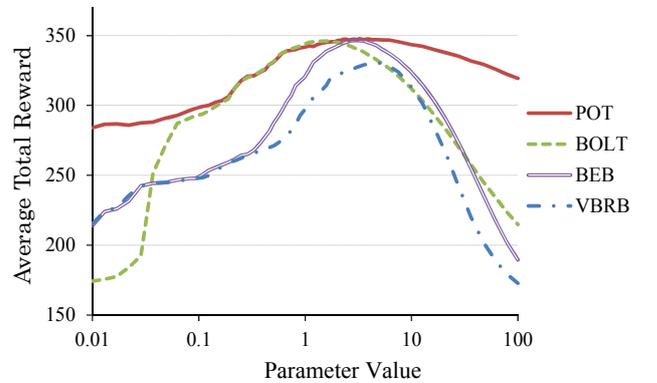

**Figure 3. Performance with no prior knowledge**

Figure 3 shows the average total reward versus the parameter value. The total rewards shown in the figure are the average based on $10^5$ runs, which made the standard error negligible (it was at most ±0.4). The results indicate that POT maintained a higher level of performance for a wider range of parameters than the other algorithms. This is exactly what was predicted in the previous sections. The existing algorisms did not work well with large parameter values because those values made the agent much too optimistic. On the other hand, they behaved poorly with small parameter values since these values let the agent be not optimistic and only exploit its current knowledge. In contrast, POT worked well even with large or small parameter values because POT adaptively changed the degree of optimism based on the information as to true MDP. In terms of the range of parameter values, where the agent achieved more than 300 rewards, POT turned out to be the best, followed by BOLT and BEB, and VBRB was the worst. We should note that the parameters' theoretical meanings differ for each algorithm (e.g., $2H^2$ for BEB, $H$ for BOLT, and $H\lambda$ for POT). Hence, comparing the results in the same scale as in figure 3 is *an abuse of notation* from a theoretical point of view. However, from a practical point of view, the parameters' theoretical meanings are almost irrelevant here. Notice that no peak of the curve in figure 3 corresponds to theoretical values of the parameters. Based on this practical standpoint, we treated all the parameters equally as one arbitrarily adjustable parameter without any meaning for the pragmatically important comparison.

**Table 1. Maximum performance with no prior knowledge**

| # | Algorithm | Parameter | Average | 90% | 10% |
|---|---|---|---|---|---|
| 1 | POT | $\beta$=3.2 | 347.5±0.1 | 386.4 | 309.8 |
| 2 | BOLT | $\eta$=1.4 | 345.7±0.1 | 385.6 | 306.8 |
| 3 | BEB | $\beta$=2.5 | 342.3±0.1 | 383.1 | 302.0 |
| 4 | MBIE-EB | $\beta$=2.5 | 336.5±0.1 | 374.6 | 298.6 |
| 5 | VBRB | $\beta p$=4.9 | 326.4±0.1 | 374.5 | 278.6 |

Table 1 summarizes the different algorithms' performances with optimal parameter settings. The algorithms are ordered in the table from highest to lowest performance. The results are based on $10^5$ runs and the standard errors are presented along with the average total rewards. The parameter of each algorithm is optimized in the same way that was adopted by previous work. Indeed, the estimated optimal parameter for MBIE-EB and BEB is the same value reported by Kolter et al. [8]. We can see that POT worked better than others and that the difference of the average total rewards tended to be ascribed to 10% value. All the algorithms in the table have the potential to obtain 370+ in several trials (see the 90% values), but only a few could assure near 310 total rewards with high probability (see the 10% values). We will discuss the reason for the relationship among the algorithms' performances in figure 3 and table 1, together with the results for misleading priors later on.

Now that we understand POT's great performance in the case of no prior knowledge, it is time to discuss its performance when we have informative knowledge regarding the dynamics. To account for this situation, we constructed informative priors by updating the uniform prior containing small amounts of information with ideal observations that are the multiple of true probabilities and the prior's sizes. The similar way to create informative prior was used in [10].

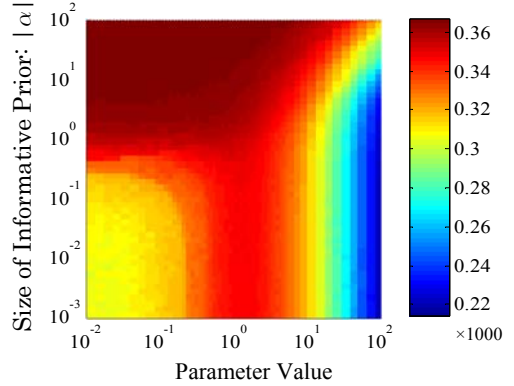

Figure 4. BOLT's performance with informative prior

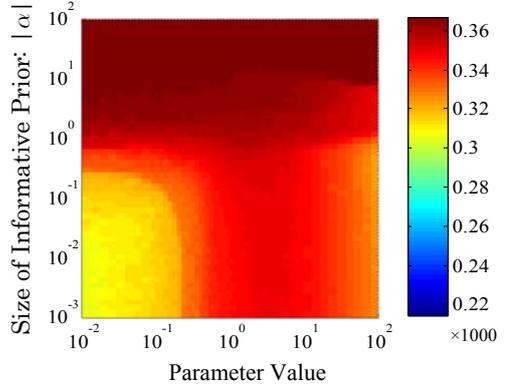

Figure 5. POT's performance with informative prior

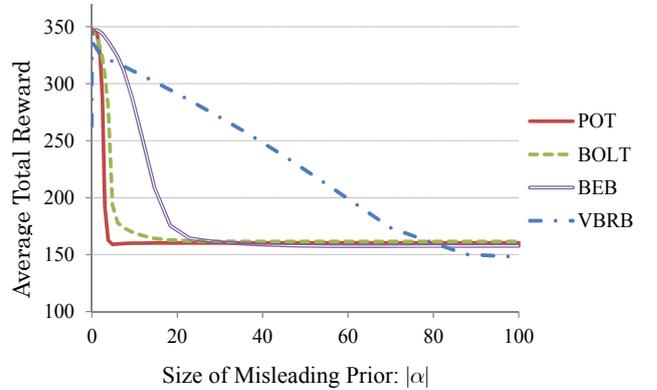

Figure 6. Performance with misspecified priors

Figure 4 and figure 5 respectively show the $10^5$ runs' average total rewards of BOLT and POT with the different degree of informative priors and their parameters. The result of BOLT is shown here as a representative of the existing algorithms (the results of the algorithms except POT turned out to be almost the same for this setting). Each figure respectively indicates that BOLT and POT can effectively utilize informative priors. That is, their average total rewards went up as the degree of information increased and they almost reached the optimal total reward (around 367.7). More importantly, the comparison of figure 4 and figure 5 tells us that unlike BOLT, POT earned at least 300 in all settings. Also, we should report that with optimal parameter settings, POT achieved the average total reward more than 350 with the prior size 0.035 while BOLT did so with the size 0.33. This implies that POT would have a superior ability to utilize a small amount of informative knowledge. Putting together the results thus far, POT could perform better than the existing algorithms when prior knowledge was either informative or not assigned to the agent. This intuitively makes sense, as POT adaptively controls the degree of its optimism and is greedier than the other algorithms.

Finally, we discuss the algorithms' ability to handle misspecified prior information. Here, misspecified prior is defined as the uniform prior with a non-small amount of information. This prior is considered to be misspecified because true transition probabilities are not uniform, and the same concept was used in [8].

Figure 6 shows the average total reward versus the degree of misspecified prior based on $10^5$ runs. As it can be seen by comparing figure 6 to figure 3 and table 1, the algorithms' ranks in terms of ability to handle misspecified priors were opposite to their ranks with non-misleading priors. For example, VBRB seems to be the best in this situation, while it was the worst in other settings. This is because VBRB, a PAC-MDP algorithm derived in Bayesian setting, could avoid being misled by misspecified priors at the expense of aiming for Bayesian optimality. On the other hand, POT, BOLT and BEB share the goal to approximate Bayesian optimal behavior. Hence, they were misled by the incorrect knowledge to the similar extent, but worked better than the PAC-MDP algorithm when the prior was reasonable. In turn, the difference between the performances of POT, BOLT and BEB can be explained by the inequality of their greediness and sample complexity. As we discussed in the previous section, POT is greedier and has lower sample complexity than the others. These theoretical properties naturally explain the results that POT worked best when the prior was reasonable and worst when the prior was misspecified.

## 5. CONCLUSION

In this paper, we introduced a new algorithm called POT, with which an agent can be greedier than with existing algorithms and perform well with a very wide range of parameter values. We derived POT by letting the agent utilize not only Bayesian optimal reasoning but also the information of potentially true MDP. More concretely, an agent with POT adaptively changes the degree of optimism as it learns where a true MDP potentially lies. With a larger than optimal parameter value, the existing algorithms usually maintain too much optimism and over explore. On the other hand, with a smaller than optimal parameter value, the existing algorithms are not optimistic enough and become stuck into a sub-optimal state. With POT, we naturally solved this issue by letting an agent have adaptive degrees of optimism. To do so, we relaxed the requirement placed by *optimism in the face of uncertainty* principle.

A consequence of relaxing the condition of the optimism was that POT does not guarantee standard Bayesian RL behavior, but Probably Upper Bounded belief-based Bayesian RL (PUB Bayesian RL) behavior. Unlike other existing approximation methods of Bayesian RL, PUB Bayesian RL does not use any myopic heuristics nor omit the search for any state-action space. Instead, PUB Bayesian RL limits possible belief-evolutions to be thought ahead in accordance with probability theory. Therefore, PUB Bayesian RL is optimal with high probability if the assigned information is not largely misleading, the condition of which also holds true for standard Bayesian RL. The difference between standard and PUB Bayesian RL was that PUB Bayesian RL is greedier than standard Bayesian RL as it exploits additional current knowledge to limit explorations. The concept of this alternative optimal behavior allowed POT to have both a lower sample complexity and the ability to explore environments more greedily than the previous algorithms.

We demonstrated the above points in the standard chain problem. As predicted, POT outperformed other algorithms when the prior distribution was not greatly misspecified. In that case, POT achieved the highest average total reward with the optimal parameter setting and also showed much lower parameter sensitivity compared to others. On the other hand, the limitation of POT was shown to be the inability to handle misspecified priors. But, we also demonstrated that the exact same drawback exists for the near Bayes-optimal algorithms when compared to a PAC-MDP algorithm.

The disadvantage of the near Bayes-optimal algorithms comes from their greater greediness than PAC-MDP, and the same can be said for POT. But, of course, the greedier behaviors have distinct advantages and we also confirmed this point in the experiment. Therefore, we can think of the selection of an algorithm, from PAC-MDP to PUB or standard Bayesian optimal algorithms, as the choice for a preferable greediness level. The preference regarding the degree of greediness should differ for different tasks and confidence levels of prior knowledge. For example, if we use a small prior or are very confident about a prior with a large bias, we may ought to choose a greedier algorithm. In this sense, our introduction of POT and the concept of PUB Bayesian RL will contribute to giving us not only a robust algorithm for varying parameter values, but also a choice to select an algorithm with a new level of greediness.

Future work includes using the concept of PUB Bayesian RL to derive other algorithms. For instance, it is interesting to see how existing algorithms like BEB or Sparse Sampling will work if they are modified with the concept of PUB Bayesian RL. Another future work would be to test POT with a variety of experiments. Because the chain problem does not largely penalize over-exploration, we may find another advantage of POT's adaptive optimism in other problems that disfavor over-exploration more.

## 6. ACKNOWLEDGMENTS

We would like to thank the anonymous reviewers for their helpful comments.